\documentclass[letterpaper, 10 pt, conference]{ieeeconf}  

\IEEEoverridecommandlockouts                              

\usepackage[T1]{fontenc}

\usepackage[space, compress, sort]{cite}

\usepackage{amssymb} 
\usepackage{amsfonts}
\usepackage{amsthm}
\usepackage{wrapfig}
\usepackage[ruled,vlined]{algorithm2e}
\usepackage{mathtools}
\usepackage{tabularx}
\usepackage{graphicx}
\usepackage{enumerate}
\usepackage{float}
\usepackage{url}
\usepackage{verbatim}
\usepackage{booktabs} 
\usepackage{multirow}
\usepackage{caption}
\usepackage{hyperref}
\usepackage{xcolor}
\usepackage{bm}
\usepackage{subfig}
\usepackage{algpseudocode}

\SetKwRepeat{Do}{do}{while}

\usepackage{xcolor}

\title{\LARGE \bf
CLAW: A Vision-Language-Action Framework for Weight-Aware Robotic Grasping
}

\author{Zijian An$^{1\star}$, Ran Yang$^{2\star}$, Yiming Feng$^{2}$, and Lifeng Zhou$^{1\dagger}$
\thanks{$^{1}$Zijian An and Lifeng Zhou are with the Department of Electrical and Computer Engineering,
        Drexel University, 3141 Chestnut St, Philadelphia, PA 19104, USA
        {\tt\small \{za382, lz457\}@drexel.edu}}%
\thanks{$^{2}$Ran Yang and Yiming Feng are with Virginia Seafood Agricultural Research and Extension Center, and Department of BiologicaCASEl Systems Engineering, Virginia Tech, 15 Rudd Ln, Hampton, VA, 23669, USA
        {\tt\small \{ryang17,yimingfeng\}@vt.edu}}%
\thanks{$\star$ Equally contributed}%
\thanks{$\dagger$ Corresponding author}%
 }

\begin{document}

\maketitle
\thispagestyle{empty}
\pagestyle{empty}

\begin{abstract}
Vision-language-action (VLA) models have recently emerged as a promising paradigm for robotic control, enabling end-to-end policies that ground natural language instructions into visuomotor actions. However, current VLAs often struggle to satisfy precise task constraints, such as stopping based on numeric thresholds, since their observation-to-action mappings are implicitly shaped by training data and lack explicit mechanisms for condition monitoring. 
In this work, we propose \textbf{CLAW} (CLIP-Language-Action for Weight), a framework that decouples condition evaluation from action generation. CLAW leverages a fine-tuned CLIP model as a lightweight prompt generator, which continuously monitors the digital readout of a scale and produces discrete directives based on task-specific weight thresholds. These prompts are then consumed by $\pi_0$, a flow-based VLA policy, which integrates the prompts with multi-view camera observations to produce continuous robot actions. This design enables CLAW to combine symbolic weight reasoning with high-frequency visuomotor control.
We validate CLAW on three experimental setups: single-object grasping and mixed-object tasks requiring dual-arm manipulation. Across all conditions, CLAW reliably executes weight-aware behaviors and outperforms both raw-$\pi_0$ and fine-tuned $\pi_0$ models. \href{https://youtu.be/MuMYj2QgReI}{A video} of our paper is available online.

\end{abstract}

\section{INTRODUCTION}
\label{introduction}
Vision--language--action (VLA) models have recently emerged as a powerful paradigm for robotic control. By grounding natural language instructions into visual observations, VLAs enable robots to perform a wide range of everyday tasks such as folding clothes \cite{black2024pi_0}, wiping surfaces \cite{ahn2024vader}, or fetching objects \cite{kim2024openvla}. These models demonstrate impressive generalization in open-ended environments, where explicit task-specific programming is complex.
Recent progress has been driven by diverse architectural designs. For instance, RT-2 \cite{zitkovich2023rt}, OpenVLA \cite{kim2024openvla}, and $\pi_0$-FAST \cite{pertsch2025fast} discretize robot actions into tokens and cast control as a next-token prediction problem, enabling scalable training on large-scale demonstration datasets. $\pi_0$ \cite{black2024pi_0}, in contrast,  adopts a flow-matching action head that models a continuous distribution over motor commands, allowing it to generate smooth and high-frequency control signals. Similarly, diffusion-based VLAs \cite{shukor2025smolvla, guo2025smovla, wen2025diffusionvla,liang2025discrete} represent yet another direction, where actions are generated through iterative denoising, thereby capturing multimodal distributions over feasible trajectories. These approaches highlight the flexibility of the VLA framework, showing that visual grounding and language conditioning can be combined with different generative mechanisms to produce robot behavior. 
However, despite their versatility, existing VLA systems often lack the ability to carry out precise and rigorously constrained actions. For example, when a task requires continuous monitoring of sensor feedback such as deciding when to stop grasping based on the exact weight of objects, current VLAs struggle to enforce such fine-grained control, as their action generation remains largely semantic and high-level.

All VLA models incorporate a vision-language model (VLM) component that provides the semantic grounding between visual inputs and natural language instructions. VLMs have also undergone rapid independent evolution in recent years. Early models, such as CLIP \cite{radford2021learning}, focused on large-scale contrastive training, aligning images and text in a shared embedding space, and enabling strong zero-shot recognition and retrieval. Building on this foundation, models like BLIP \cite{li2022blip}, SigLIP \cite{zhai2023sigmoid}, LLaVA \cite{liu2023visual, liu2024improved}, and OpenFlamingo \cite{awadalla2023openflamingo} extended VLMs beyond alignment, equipping them with generative and conversational capabilities for captioning, visual question answering, and instruction following. More recently, DAM-3B~\cite{lian2025describe} represents a further step forward, demonstrating the ability to produce fine-grained, structured descriptions of visual scenes that include numerical indicators, spatial relations, and symbolic cues. This trajectory illustrates how VLMs have progressed from simple cross-modal alignment toward increasingly rich semantic grounding and structured perception. 

In standard VLA architectures, however, the role of the embedded VLM is primarily to facilitate fluent action generation, focusing on producing plausible visuomotor trajectories rather than enforcing precise task constraints. This observation motivates our work: we argue that VLA performance can be enhanced by introducing an additional, lightweight, and task-specialized VLM that serves as an external corrective module. Such a design highlights fine-grained cues, for example, digits on a scale or localized scene attributes, that may otherwise be overlooked by the main VLM, and upgrades the objective from merely ``completing an action'' to truly ``completing the task.'' In doing so, the augmented VLA system gains the ability to achieve more flexible and quantitatively accurate robotic control. 

Building on this idea, we present CLAW (CLIP–Language–Action for Weight), a framework that enhances the standard VLA architecture with an additional task-specific VLM. CLAW employs a fine-tuned CLIP model as a lightweight perception module that monitors the digital display of a weight scale and converts it into symbolic language prompts. These prompts are then combined with multi-view camera observations of the robot’s state and passed to $\pi_0$ for generating accurate robotic motions. By explicitly introducing this prompt generation stage, CLAW forms a closed loop from perception to language to action. This design enables precise, weight-aware grasping that addresses limitations of existing end-to-end VLAs.  
Since prompt evaluation is invoked in real-time during manipulation, the task-specific VLM must operate at a high frequency. This requirement rules out large generative VLMs, which are often too slow for continuous control. Moreover, the VLM in our framework only needs to provide binary weight-based prompts (\texttt{continue} or \texttt{stop}), rather than producing rich captions or detailed scene descriptions. To meet these constraints, we adopt CLIP~\cite{radford2021learning} as a lightweight vision–language model and fine-tune it on a custom dataset of scale readings. This choice ensures fast inference and robust prompt generation, while avoiding the overhead of larger models such as DAM-3B~\cite{lian2025describe}. We evaluate CLAW on three representative scenarios: candy picking, garlic picking, and a mixed-task setup involving both objects. These experiments demonstrate the robustness of CLAW across different objects and environments. The results show that CLAW not only performs reliable grasping under varying conditions but also consistently respects weight constraints. 

Our work makes the following contributions:
\begin{itemize}
    \item We introduce CLAW, a framework that augments standard VLA architectures with an additional, lightweight VLM for explicit condition monitoring, enabling weight-aware robotic manipulation.  
    \item We design a fine-tuning procedure for CLIP that allows it to serve as a reliable prompt generator, translating scale readings into VLA understandable language prompts.
    \item We finetune $\pi_0$ with prompt supervision, ensuring that it can effectively integrate CLIP-generated prompts with multi-view observations to produce precise actions.  
    \item We evaluate CLAW on three representative scenarios and demonstrate that it achieves reliable grasping under varying conditions, respects weight constraints, and outperforms naive baselines, raw $\pi_0$, and fine-tuned $\pi_0$.  
\end{itemize}

\section{RELATED WORK}
\label{related work}
\subsection{VLMs and CLIP}
\textbf{VLMs} aim to connect visual perception with natural language, enabling machines to understand and communicate about visual content. Early developments in this area primarily focused on aligning images and text, where models were trained to determine whether an image matched a given textual description. Representative approaches, such as CLIP \cite{radford2021learning}, have demonstrated that large-scale contrastive pre-training on image–text pairs can produce transferable visual representations, supporting zero-shot recognition and classification across diverse tasks. This line of work established VLMs as powerful perception modules that can bridge visual inputs with high-level semantic concepts.

More recent advances have moved beyond alignment toward fine-grained and generative capabilities. Instead of merely determining correspondence between images and textual labels, new models can produce detailed and context-sensitive descriptions of objects, scenes, and even spatiotemporal dynamics in videos. Systems such as DAM-3B \cite{lian2025describe} illustrate this shift, showing that VLMs can localize regions of interest and generate nuanced, multi-sentence captions grounded in both global context and local details. These developments underscore the increasing capability of VLMs not only to classify and align, but also to interpret, describe, and reason about complex visual environments.

\textbf{CLIP} learns to align images with natural language descriptions in a shared embedding space. Specifically, CLIP jointly optimizes an image encoder $f_\theta(\textbf{I})$ and a text encoder $g_\phi(\textbf{T})$. Given a minibatch of $N$ image--text pairs $\{(\textbf{I}_i, \textbf{T}_i)\}_{i=1}^N$, both encoders map their inputs into a common $d$-dimensional space:  
$$
\textbf{v}_i = f_\theta(\textbf{I}_i), \quad \textbf{u}_i = g_\phi(\textbf{T}_i),
$$  
followed by normalization  
$ \hat{\textbf{v}}_i = \textbf{v}_i / \|\textbf{v}_i\|, \quad \hat{\textbf{u}}_i = \textbf{u}_i / \|\textbf{u}_i\| $.  
The similarity between image and text embeddings is measured by cosine similarity  
$$
s_{ij} = \hat{\textbf{v}}_i^\top \hat{\textbf{u}}_j.
$$  
CLIP is trained with a symmetric cross-entropy loss over all possible pairings within the batch:  
$$
\mathcal{L} = \frac{1}{2}\Bigg( \frac{1}{N}\sum_{i=1}^N \mathrm{CE}(s_{i,:}, i) + \frac{1}{N}\sum_{j=1}^N \mathrm{CE}(s_{:,j}, j) \Bigg),
$$  
where $\mathrm{CE}(\cdot)$ denotes the cross-entropy and the target index corresponds to the true pairing. Through large-scale pre-training on 400M image--text pairs, CLIP learns general-purpose visual representations that can be directly applied to downstream tasks in a zero-shot manner by embedding candidate class names as text prompts and selecting the most similar one.  

\subsection{VLA and $\pi_0$}
\textbf{VLA Models.} 
Recent advances in multimodal learning have led to the development of VLA models, which extend VLMs by directly producing robot control actions. 
A typical VLA takes as input visual observations $\textbf{o}_t$ (e.g., camera images) together with natural language instructions $l$, and outputs robot actions $\textbf{a}_t$ at each timestep. 
Formally, the model parameterizes a conditional distribution 
\begin{equation}
p_\theta(\textbf{a}_t \mid \textbf{o}_t, l, \textbf{a}_{<t}),
\label{eq:vla_principle}
\end{equation}

\begin{figure*}[h]
    \centering
    \vspace{2mm}
    \includegraphics[width=\linewidth]{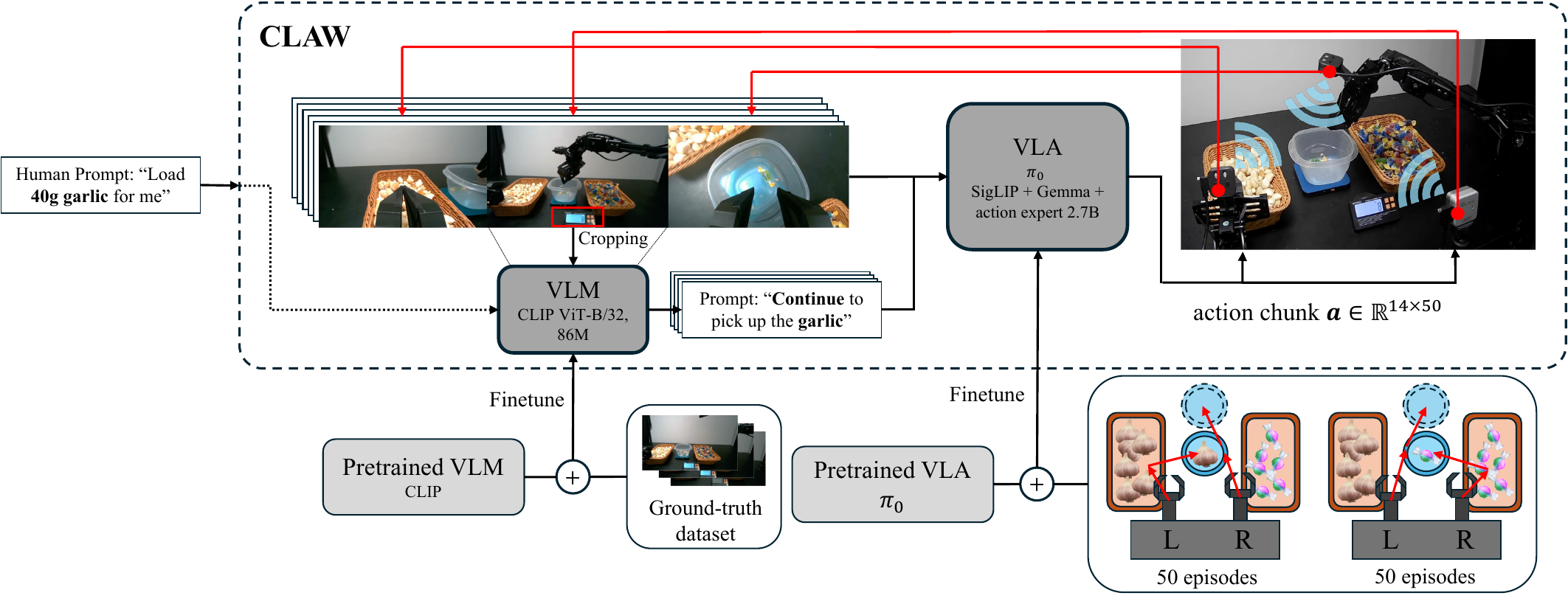}
    \vspace{-2pt}
    \caption{Architecture of CLAW. A human instruction is provided to the fine-tuned CLIP module, which monitors the scale and generates a symbolic prompt indicating the target object and whether the weight goal has been reached. The fine-tuned $\pi_0$ VLA then takes this prompt along with multi-view observations as input and produces action chunks that drive the robot to execute the task.}
    \label{fig:framework}
\end{figure*}

where actions are treated as tokens in the same vocabulary as natural language. 
The training objective is usually formulated as a standard next-token prediction loss: given paired trajectories $\{(\textbf{o}_t, l, \textbf{a}_t)\}$, the model minimizes the cross-entropy between predicted and ground-truth tokens. 
This formulation allows VLAs to inherit the strong semantic grounding of pretrained VLM backbones, while adapting them to visuomotor control. 
Compared with modular pipelines that separately handle perception, planning, and control, VLA models offer a unified, end-to-end approach that can scale with large datasets. 
Importantly, different architectures instantiate Eq.~\ref{eq:vla_principle} in distinct ways: some adopt an autoregressive next-token prediction scheme, others employ diffusion-based denoising of action trajectories, and yet others rely on flow-matching dynamics to generate continuous control signals, as introduced in Section \ref{introduction}. 
Moreover, there also exist approaches that distill high-level policies into LLMs \cite{ahn2022can}, where the language model is trained to directly decide which primitive skill or low-level controller to invoke, effectively transferring expert policy knowledge into a linguistic reasoning framework. Another line of work follows a hybrid skill-based design \cite{driess2023palm, zhao2025unveiling}, in which the VLA outputs symbolic subgoals or skill tokens that are then executed by dedicated motion controllers. These approaches differ from next-token, diffusion, or flow-matching VLAs in that they explicitly separate high-level decision-making from low-level control, pursuing a modular rather than fully end-to-end integration of language and action.

\textbf{The $\pi_0$ Model} is a representative example of end-to-end VLAs. We adopt $\pi_0$ as the underlying VLA policy due to its ability to generate continuous high-frequency control via flow matching, offering both precision and efficiency compared to token-based or diffusion-based alternatives. Additionally, it is the state-of-the-art VLA model that provides a balance of performance, stability, and compatibility with language-conditioned prompting in our setting.  

The $\pi_0$ builds on a pretrained VLM backbone for multimodal encoding and augments it with a flow-based policy head that generates continuous robot actions. 
Instead of discretizing actions into tokens, $\pi_0$ models a continuous distribution over actions by learning a time-dependent velocity field $\textbf{v}_\theta(\textbf{x},t)$ that transports samples from a simple prior (e.g., Gaussian noise) to the empirical action distribution. 
Concretely, given an initial noise sample $\textbf{x}(0) \sim p_0$ and a target action $\textbf{a} \sim p_{\text{data}}$, the flow dynamics are defined by the ODE
$$
\frac{d \textbf{x}(t)}{dt} = \textbf{v}_\theta(\textbf{x}(t), t, \textbf{o}_t, l),
$$
with the terminal condition $\textbf{x}(1) \approx \textbf{a}$. 
The training objective, known as the flow matching loss, minimizes the squared distance between the learned velocity field and the ideal displacement toward the target:
$$
\mathcal{L}_{\text{flow}} = \mathbb{E}_{t \sim U(0,1), \, (\textbf{o}_t, l, \textbf{a})} \left[ \, \big\| \textbf{v}_\theta(\textbf{x}(t), t, \textbf{o}_t, l) - (\textbf{a} - \textbf{x}(t)) \big\|^2 \, \right].
$$
At inference time, robot actions are generated by integrating the learned flow field from noise to data space, yielding smooth and high-frequency control signals. 
This design enables $\pi_0$ to operate at up to 50Hz, producing precise low-level commands for manipulation. 
Moreover, $\pi_0$ is trained on diverse cross-embodiment datasets spanning thousands of hours of demonstrations, which equips it with strong generalization ability across robots and tasks.

While VLA models such as $\pi_0$ provide a unified framework that maps observations and prompts to continuous action sequences, their decision-making is ultimately shaped by the statistical correlations present in the training data. In practice, this means that the observation stream $\textbf{o}_t$ is processed holistically, and the model implicitly decides which aspects of the input to attend to during action generation. As a consequence, VLAs often struggle to satisfy explicit human requirements that demand prioritizing specific regions, elements, or features in the visual input. For example, suppose a task requires the robot to continuously monitor a particular location on a scale or track the state of a small object. In that case, a purely end-to-end VLA may fail to allocate sufficient attention and to adjust its behavior accordingly. 


\section{APPROACH}
\subsection{CLAW Pipeline}
Our proposed CLAW framework is shown in Figure \ref{fig:framework}. The robot executes weight-conditioned manipulation tasks by coupling a lightweight VLM, CLIP, with $\pi_0$. At the beginning of each episode, a human-specified task instruction is provided, such as ``\texttt{load 50 g candy for me.}'' This instruction is passed to the CLIP module, which continuously observes images of the scale at a fixed frequency. Based on the current visual reading, CLIP evaluates whether the measured weight has reached the specified threshold. Formally, CLIP parameterizes a conditional distribution
$$
p_\phi(m_t \mid \textbf{o}_t^{\text{scale}}, \, l),
$$
where $\textbf{o}_t^{\text{scale}}$ denotes the image of the scale at time $t$, $l$ is the human instruction, and $m_t \in \{\texttt{continue}, \texttt{stop}\}$ represents the generated prompt. 
These prompts are subsequently fed into the $\pi_0$ policy, which receives them alongside multi-view observations $\textbf{o}_t^{\text{scene}}$ from three cameras providing different perspectives of the workspace. Conditioned on both the visual context and the prompt, $\pi_0$ produces continuous low-level actions according to
$$
p_\theta(\textbf{a}_t \mid \textbf{o}_t^{\text{scene}}, \, m_t),
$$
where $\textbf{a}_t$ denotes the robot control action at time $t$. In this way, CLAW integrates symbolic weight monitoring with visuomotor control: CLIP provides explicit guidance based on task-specific thresholds, and $\pi_0$ translates this guidance into precise motor commands, enabling reliable execution of weight-aware manipulation tasks.

\subsection{Fine-Tuning CLIP for Weight-Based Prompting}
We first evaluated the zero-shot performance of CLIP on the task of interpreting scale readings. While CLIP is capable of recognizing digits and associating them with textual labels, we found that its zero-shot accuracy for numerical comparison was unsatisfactory. 
This motivated us to fine-tune CLIP specifically for weight-conditioned prompting.
\begin{figure}[h]
    \centering
    \vspace{2mm}
    \includegraphics[width=\linewidth]{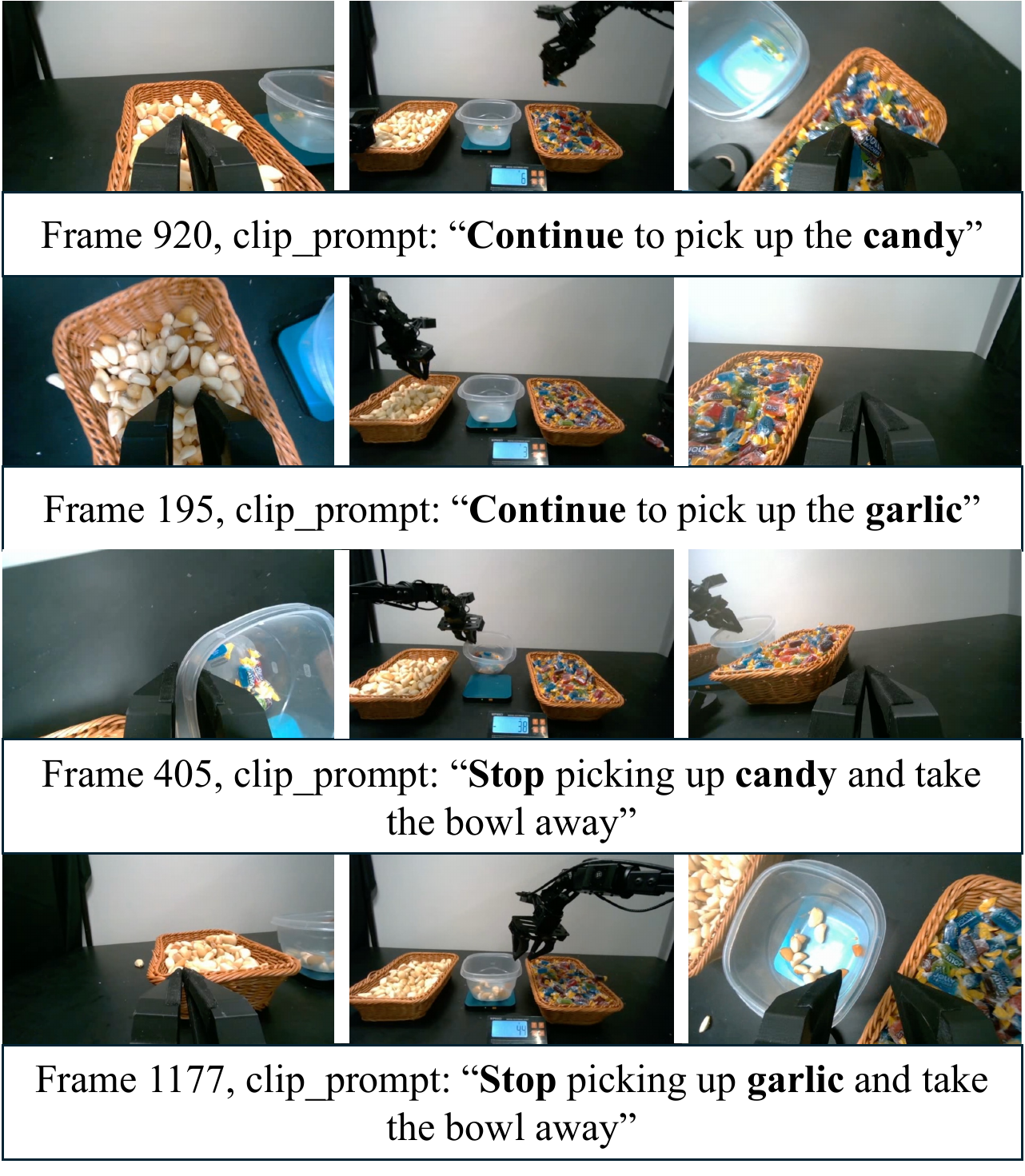}
    \caption{The outputs of CLIP under different observations.}
    \label{fig:clip_prompt}
\end{figure}

To construct a training dataset, we randomly sampled $2000$ images from the fixed camera that observes the scale. For each image, we cropped the region corresponding to the digital display of the scale. Given a ground-truth weight value $w^\ast$ read from the image, we paired the crop with $N$ synthetic task instructions of the form ``\texttt{load $k$ g target for me.}'' where $k \in \{1, 2, \dots, N\}$. Each $(\text{image}, \text{instruction})$ pair was assigned a binary label: 
$$
y = 
\begin{cases}
\texttt{continue}, & \text{if } k < w^\ast, \\
\texttt{stop}, & \text{if } k \geq w^\ast.
\end{cases}
$$
This procedure yielded $2000 N $ labeled training samples.

We then fine-tuned CLIP on this dataset to optimize the classification loss between the predicted label and the ground-truth $y$. After fine-tuning, the adapted CLIP model is capable of robustly mapping each scale image and task instruction to a discrete prompt $m_t \in \{\texttt{continue}, \texttt{stop}\}$, which is subsequently provided to $\pi_0$ for action generation. The finetuned CLIP model can continuously interpret scale readings and reason in real time about which action should be executed next, as illustrated in Figure \ref{fig:interruption}.

\begin{figure*}
    \centering
    \vspace{2mm}
    \includegraphics[width=.8\linewidth]{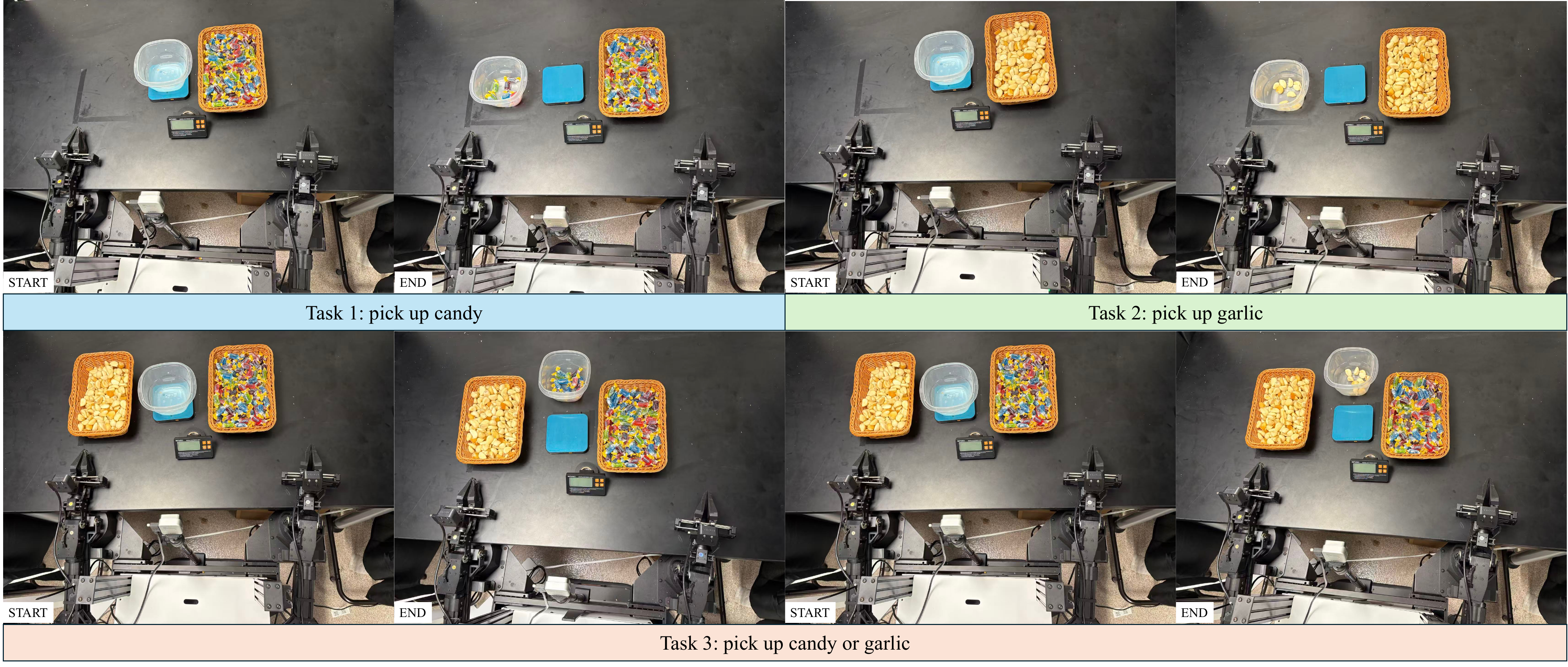}
    \caption{Configurations for different tasks. In single-object settings, CLAW achieves weight-specified grasping, while in multi-object settings, it enables grasping of a specified object at a specified weight.}
    \label{fig:tasks}
\end{figure*}
\begin{table*}[thb]
    \centering
    \caption{Success rate comparison.}
    \vspace{-1mm}
    \label{tab:entropy_comparison}
    \begin{tabular}{lcccccc}
    \toprule
    \textbf{Task} & \multicolumn{2}{c}{\textbf{Raw-$\pi_0$}} & \multicolumn{2}{c}{\textbf{Fine-tuned $\pi_0$}}& \multicolumn{2}{c}{\textbf{CLAW}} \\
    \midrule
    &\textit{Action}&\textit{Stop point}&\textit{Action}&\textit{Stop point}&\textit{Action}&\textit{Stop point} \\
    \midrule
    Pick up 20g candy & 0.35& 0.05& 1.00&0.15& 1.00& 1.00 \\
    Pick up 30g candy & 0.25& 0.00& 1.00&0.35& 1.00& 1.00 \\
    Pick up 40g candy & 0.30& 0.00& 1.00&0.20& 1.00& 1.00 \\
    Pick up 20g garlic & 0.20& 0.00& 1.00&0.15& 1.00& 1.00 \\
    Pick up 30g garlic & 0.20& 0.00& 1.00&0.10& 1.00& 1.00 \\
    Pick up 40g garlic & 0.15& 0.00& 1.00&0.00& 1.00& 1.00 \\
    \bottomrule
    \end{tabular}
    \vspace{1mm}
    \label{tab}
\end{table*}
\subsection{Fine-Tuning $\pi_0$ with Prompt Supervision}
To adapt $\pi_0$ for our weight-aware manipulation setting, we collected $50$ demonstration episodes for each task. Each episode contained two phases: (i) grasping the specified target object, and (ii) terminating the grasp and removing the bowl. During data collection, CLIP was not involved. Instead, we manually annotated each frame with a \texttt{clip\_prompt} label to simulate the role of CLIP, as illustrated in Figure~\ref{fig:clip_prompt}. All frames corresponding to the grasping phase were labeled with the prompt ``\texttt{Continue to pick up the target.}'' whereas frames corresponding to the termination phase were labeled with ``\texttt{Stop picking up and take the bowl away.}'' 

Given observation frames $\textbf{o}_t^{\text{scene}}$ from three cameras, the annotated prompt $m_t$, and ground-truth robot actions $\textbf{a}_t$, we fine-tuned $\pi_0$ by minimizing the flow-matching loss conditioned on both the visual input and the prompt:
$$
\mathcal{L}_{\pi_0} = \mathbb{E}_{(\textbf{o}_t^{\text{scene}}, m_t, \textbf{a}_t)} 
\Big[ \, \| \textbf{v}_\theta(\textbf{x}(t), t, \textbf{o}_t^{\text{scene}}, m_t) - (\textbf{a}_t - \textbf{x}(t)) \|^2 \, \Big],
$$
where $\textbf{v}_\theta$ denotes the learned velocity field of the flow-based policy head. 

At deployment time, the \texttt{clip\_prompt} labels are no longer manually annotated. Instead, they are provided in real time by the fine-tuned CLIP module at a fixed frequency, ensuring that $\pi_0$ receives structured guidance based on the current weight state. This design enables CLAW to combine explicit prompt supervision with end-to-end visuomotor control.

\section{EXPERIMENTS}
\begin{figure*}[th]
    \centering
    \vspace{2mm}
    \subfloat[Workflow of baseline fine-tuned $\pi_0$ when the task is loading 20g garlic.]
    {
        \includegraphics[width=0.95\linewidth]{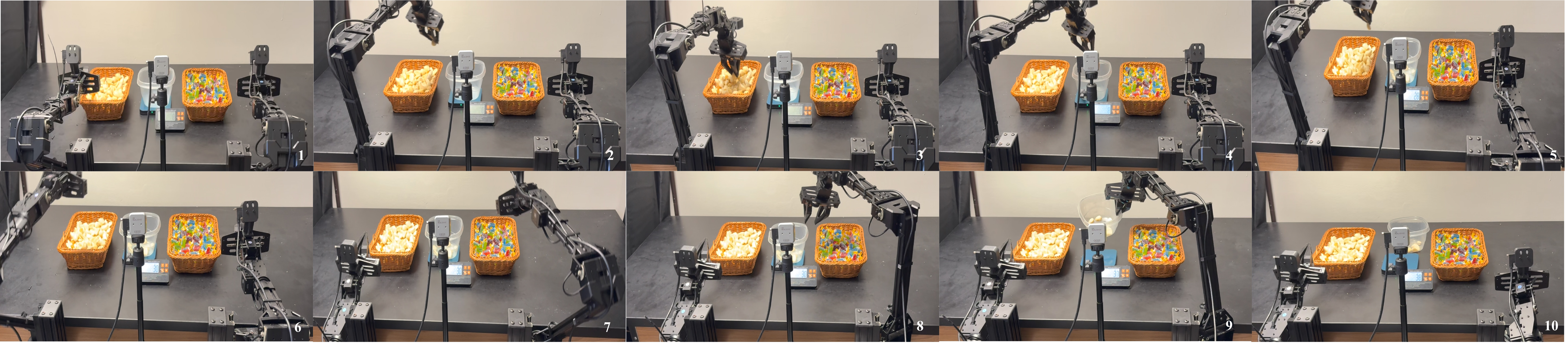}
        \label{fig:workflow:raw}
    }   
    \vspace{0.5em} 
    \subfloat[Workflow of CLAW when the task is loading 20g garlic.]
    {
        \includegraphics[width=0.95\linewidth]{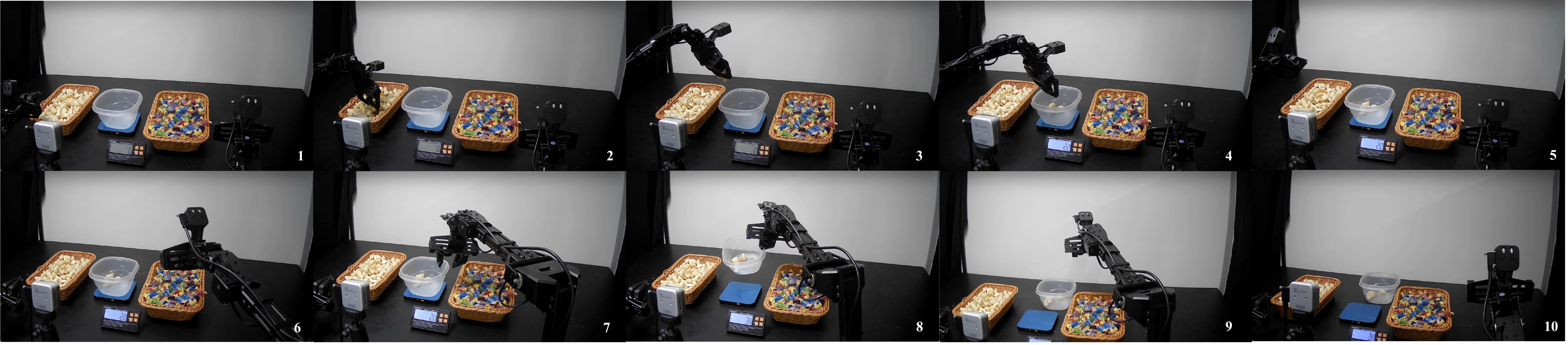}
        \label{fig:workflow:claw}
    }

    \caption{Comparison of workflows: (a) the fine-tuned $\pi_0$ does not rely on the scale reading. It tends to terminate after a random number of grasps (e.g., around seven), which may result in overshooting the desired weight (30g in this example, despite the 20g target). (b) CLAW, in contrast, monitors the scale continuously and stops immediately once the reading reaches the specified threshold of 20g.}
    \label{fig:workflow}
\end{figure*}

\subsection{Baseline: Raw $\pi_0$ and Fine-tuned $\pi_0$}

To evaluate our proposed CLAW framework, we compared it against baselines using the raw $\pi_0$ policy and fine-tuned $\pi_0$ policy without CLIP integration. In this setting, the prompt specified both the object type and the target weight (e.g., ``\texttt{pick up 20g candy.}''), but no auxiliary condition-monitoring module was provided. During data collection, we enforced termination strictly at the weight threshold: each demonstration episode consisted of grasping until the specified weight was reached, followed by termination. The experiment setup is demonstrated in Figure \ref{fig:tasks}. We collected $50$ such episodes for the candy task and fine-tuned $\pi_0$ on this dataset. We repeatedly deploy the policy under each human prompt for 20 trials, and the success rates are reported in Table~\ref{tab}. In this table, we divide the evaluation into two aspects: first, whether the robot can successfully execute the complete grasp-and-place action (the \textit{Action} columns), and second, whether it can stop at the correct weight (the \textit{Stop Point} columns).

As shown in Table~\ref{tab}, the raw $\pi_0$ policy can sometimes execute the grasping action correctly, but it is generally unable to stop at the correct time. The fine-tuned $\pi_0$ policy, after being trained on 50 demonstration episodes, reliably performs both grasping and bowl-removal actions; however, it still fails to consistently stop at the target weight. We demonstrate one example of the fine-tuned $\pi_0$ workflow in Figure \ref{fig:workflow:raw}. Although one fixed camera observed the scale throughout execution, the trained policy did not appear to use this information as a decisive feature. Instead, it reproduced behavior correlated with the number of grasps observed during training. For example, when the training demonstrations terminated after $5$-$8$ grasps (depending on the variable quantity per grasp), the deployed policy also tended to stop after $5$-$8$ attempts, albeit randomly within that range. Similarly, when the demonstrations terminated after $2$--$3$ grasps, the deployed policy reproduced this shorter horizon. Across repeated trials, the stopping point varied randomly and showed no consistent relationship to the scale reading.

In addition to this stochastic behavior, the baseline approach suffers from another fundamental limitation: each trained model is implicitly tied to a single fixed stopping weight. For example, if the ground truth demonstrations enforce termination at $50$g, the resulting policy can only reproduce that specific stopping condition and cannot adapt to other thresholds. Thus, even in principle, the raw $\pi_0$ model cannot flexibly generalize to new weight constraints, further highlighting the necessity of decoupling condition monitoring from action generation as realized in CLAW.

\subsection{Pick up Candy or Garlic}
We first evaluate CLAW on single-object grasping tasks, where the goal is to pick up either garlic or candy until a specified weight threshold is reached. The tabletop setup consists of a basket containing the target object (garlic or candy), a digital scale, and an empty bowl, as illustrated in Figure \ref{fig:tasks}. Taking the candy task as an example, CLIP is responsible for monitoring the scale and producing prompts such as ``\texttt{Continue to pick up the candy.}'' or ``\texttt{Stop picking up and take the bowl away.}'' During the fine-tuning of $\pi_0$, each frame in the demonstration dataset was manually annotated with the corresponding \texttt{clip\_prompt} label. We trained the model for $60{,}000$ steps on an H200 GPU. At inference time, $\pi_0$ operated at a control frequency of $30$Hz, generating action chunks of $50$ time steps, while CLIP updated its judgment at $20$Hz. To maintain synchronization, $\pi_0$ reads the most recent available prompt; if CLIP had not updated at a given timestep, the previous prompt was reused. 

It is worth noting that, unlike the $\pi_0$ baselines, we did not need to enforce termination at a specific weight threshold when collecting the $50$ demonstration episodes. Instead, we can collect diverse grasping and bowl-removal demonstrations, agnostic to weight threshold, object grasped, and even episodes, provided that the \texttt{clip\_prompt} labels were accurate. All grasping frames were labeled as ``\texttt{continue}'', and all bowl-removal frames as ``\texttt{stop}''. Since the baseline results indicated that the $\pi_0$ does not reliably condition on the scale reading, CLAW shifts this responsibility to CLIP. As long as CLIP generates the correct prompts, $\pi_0$ can effectively distinguish between the two commands. This design allows a single $\pi_0$ model and a single CLIP model to generalize across arbitrary weight thresholds.

We tested the system under three conditions: (i) ignoring the scale reading and forcing CLIP to always output the ``\texttt{continue}'' prompt; (ii) forcing CLIP to always output the ``\texttt{stop}'' prompt; and (iii) using CLIP to generate prompts based on the actual scale reading. In the first case, the robot continued grasping indefinitely. In the second case, the robot, upon waking from its idle state, immediately removed the bowl. In the third case, the robot successfully executed the desired behavior, picking up the bowl until the target weight was reached, and then taking it away. Interestingly, because candies are released into the bowl from a certain height, the scale readings occasionally fluctuate. When a transient spike momentarily exceeded the weight threshold, the robot exhibited an initial tendency to remove the bowl, but then returned to its grasping behavior as the reading stabilized. This behavior, although not explicitly present in the training dataset, illustrates the robustness of CLAW in handling unexpected input variations.
\begin{figure}
    \centering
    \vspace{2mm}
    \includegraphics[width=\linewidth]{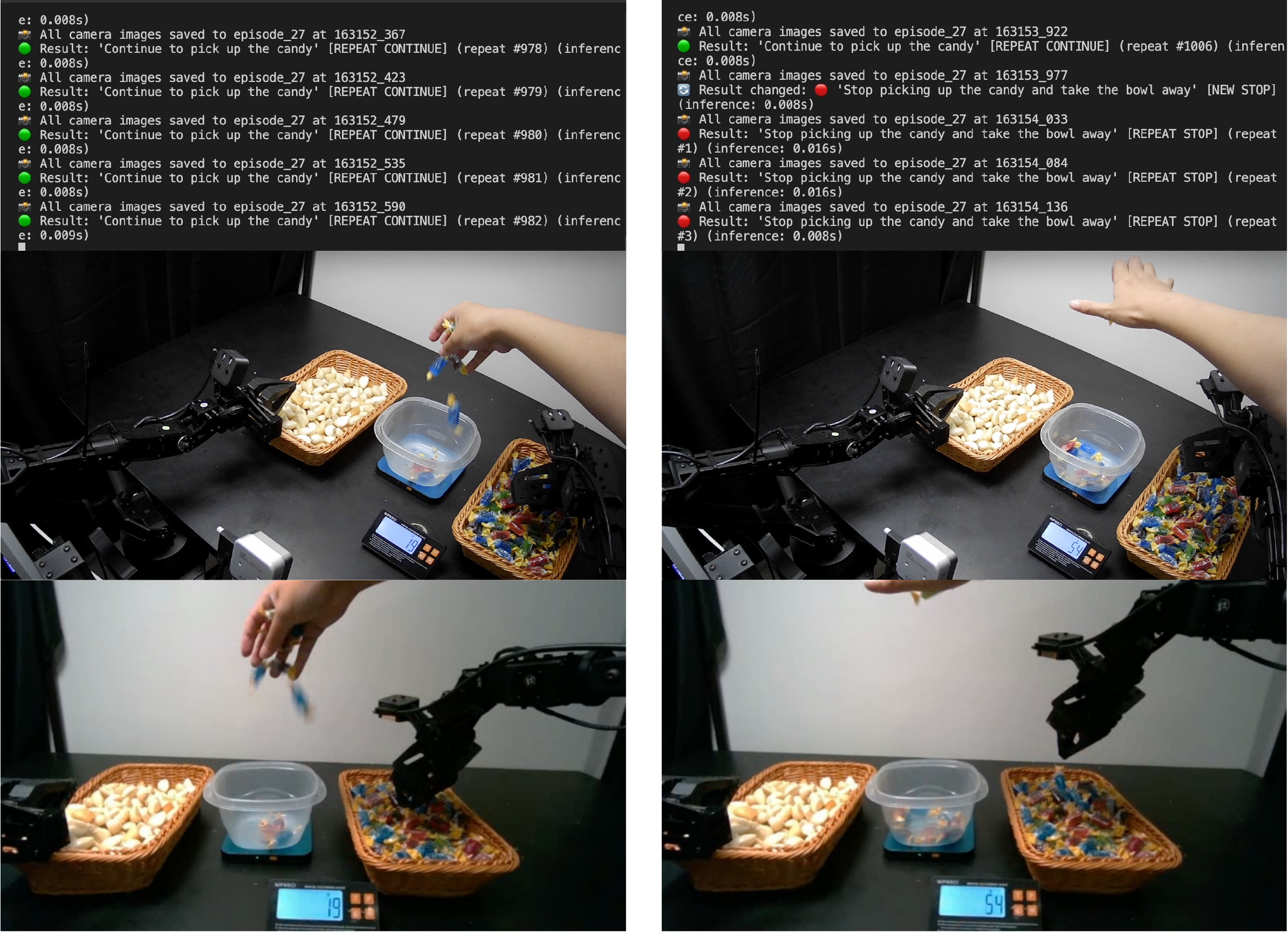}
    \caption{Robustness test during gasping candy. The weight threshold is set to 40g, and excess candy is manually added to cause the scale reading to suddenly exceed the threshold. The top terminal shows CLIP outputs. In the left figure (before dropping), CLIP reasons to continue grasping, as the scale reads 19g, which is below the 40g threshold. In the right figure (after dropping), once the scale reaches 54g, CLIP switches its reasoning to stop grasping. }
    \label{fig:interruption}
\end{figure}

\subsection{Pick up Specified Object in a Mixed Setting}
We further designed a mixed-object experiment, where a box of garlic and a box of candy were placed on the left and right sides of the table, respectively, as demonstrated in Figure \ref{fig:tasks} and Figure \ref{fig:workflow}. During fine-tuning of $\pi_0$, we collected $50$ demonstration episodes for each object, covering both grasping and bowl-removal phases. Each frame was annotated with one of four \texttt{clip\_prompt} labels: ``\texttt{Continue to pick up the candy}'', ``\texttt{Stop picking up the candy and take the bowl away}'', ``\texttt{Continue to pick up the garlic}'', and ``\texttt{Stop picking up the garlic and take the bowl away}'', as illustrated in Figure \ref{fig:clip_prompt}. To enforce role differentiation between the two arms, the right arm was assigned to grasp the candy, while the left arm removed the candy bowl. Conversely, the left arm grasped the garlic, and the right arm removed the garlic bowl. 

At deployment time, the prompt given to CLIP specified both the object type and target weight, for example ``\texttt{load 20 g candy for me.}'' CLIP then monitored the scale and produced prompts corresponding to the chosen object. The rest of the setup remained identical to the single-object experiments. We again evaluated the system under three conditions: (i) forcing CLIP to always output a ``\texttt{continue}'' prompt, which caused the designated arm to continue grasping indefinitely; (ii) forcing CLIP to always output a ``\texttt{stop}'' prompt, which immediately triggered the bowl-removal action by the opposite arm; and (iii) using CLIP to generate prompts based on the actual scale reading, in which case the robot successfully grasped the specified object until the target weight was reached and then removed the corresponding bowl. In all cases, CLAW produced the expected behavior. For example, Figure \ref{fig:workflow:claw} illustrates several characteristic motions of the robot arm observed during execution, highlighting how CLAW coordinates grasping and bowl-removal behaviors under the weight-based prompting scheme.

\textbf{Robustness test}: We conducted a disturbance test in which large amounts of extraneous items were deliberately dropped into the box during grasping, causing the scale reading to exceed the preset threshold, as shown in Figure \ref{fig:interruption}. In response, CLIP immediately output a ``\texttt{stop}'' prompt, and $\pi_0$ switched to the bowl-removal behavior. Importantly, if this disturbance occurred while one arm was still transporting the target object toward the bowl, the motion was aborted immediately and the arm retracted, while the opposite arm simultaneously began executing the removal action. This result highlights the responsiveness of CLAW to unexpected changes and its ability to coordinate dual-arm manipulation control.

\section{CONCLUSIONS AND FUTURE WORKS}
\subsection{Conclusions}
In this work, we introduced CLAW, a vision--language--action framework that integrates CLIP as a prompt generator with $\pi_0$ as a visuomotor policy to enable weight-aware robotic manipulation. Through a series of experiments on garlic and candy grasping tasks, as well as mixed-object settings with dual-arm coordination, we demonstrated that CLAW can effectively combine symbolic weight monitoring with continuous visuomotor control. 

Our findings highlight two important observations. First, $\pi_0$ exhibits high sensitivity to the prompts it receives. In the mixed-object experiments, the prompts differed only in the specification of the target object, while the remainder of the instruction remained nearly identical. Despite this high degree of similarity, $\pi_0$ was able to reliably disambiguate the prompts and execute the correct object-specific behavior. This underscores the strong conditioning effect of natural language on action generation in VLA models. Second, although CLIP is a relatively basic vision-language model, our fine-tuning procedure enabled it to function as an effective prompt generator. By accurately mapping scale readings to simple language directives, CLIP provided $\pi_0$ with reliable task guidance, ensuring that actions were triggered at the appropriate weight thresholds.

Together, these results demonstrate that CLAW achieves robust weight-conditioned robotic control by coupling a lightweight VLM with a flow-based VLA, and they suggest promising directions for future extensions of prompt-guided manipulation systems.
\subsection{Future works}
Our study opens several avenues for future research:

\textbf{Robust scale localization}: In our current setup, the scale must remain approximately fixed in position so that a cropped region consistently captures the numeric display. We observed that feeding the entire image directly into CLIP without cropping leads to reduced accuracy. A promising direction is to design an enhanced variant of CLIP that learns to automatically localize the scale within the full scene. By leveraging reinforcement learning, the model could actively search for regions containing digits or scale-like features and then perform targeted recognition, thereby removing the need for manual cropping.
    
\textbf{Generalizing stopping conditions beyond numbers}: While weight thresholds represent a natural stopping condition, many real-world tasks rely on additional modalities or features. For example, stopping criteria could be based on elapsed time, specific colors or shapes, or visual configurations that indicate task completion. Illustrative cases include terminating a task after a fixed duration, stopping cake cutting once a certain shape emerges, or recognizing success in the Tower of Hanoi puzzle when rings on each pole form the required inverted-triangle configuration. Extending CLAW with the ability to interpret such diverse cues through VLMs would enable broader applicability to complex manipulation scenarios.
    
\textbf{Multimodal integration}: Beyond visual information, incorporating additional sensory modalities could further enhance CLAW's robustness and flexibility. For instance, auditory cues (e.g., speech commands or environmental sounds) and haptic feedback (e.g., force or tactile signals from the gripper) could serve as complementary inputs for prompt generation. Integrating such multimodal signals would allow the system to better capture human intent and adapt to complex, dynamic environments where visual information alone may be insufficient.

Ultimately, the most general instantiation of CLAW should consist of two cooperating modules: a multimodal model and a VLA model. The multimodal model would determine whether the task should continue or terminate, based on heterogeneous sensory inputs, and should be capable of autonomously selecting the most relevant content within those inputs (e.g., identifying digits in an image or detecting specific frequencies in an audio signal). By decoupling this decision-making process from the VLA itself, the VLA can focus exclusively on action generation, while the multimodal module provides reliable, interpretable signals that regulate task progression.

\addtolength{\textheight}{-12cm}   









\bibliography{refs}

\end{document}